\documentclass{article}
\usepackage[nohyperref, accepted]{icml2020} 
\usepackage{microtype}
\usepackage{graphicx}
\usepackage{subfigure}
\usepackage{booktabs} 

\usepackage{hyperref}
\usepackage{color,soul}
\usepackage{graphicx}
\usepackage[utf8]{inputenc}
\usepackage{amsmath}
\usepackage{amsfonts}
\usepackage{amssymb}
\usepackage{tikz}
\usetikzlibrary{fit,positioning,quotes,arrows.meta}
\usepackage{mathtools}
\usepackage{float}

\DeclareMathOperator*{\argmax}{arg\,max}
\DeclareMathOperator{\E}{\mathbb{E}}

\DeclareMathOperator{\DKL}{\text{D}_\text{KL}}
\DeclareMathOperator{\F}{\Delta \text{F}_{\text{par}}}

\usepackage{etoolbox}

\icmltitlerunning{Hierarchical Expert Networks for Meta-Learning}

\begin{document}

\twocolumn[
\icmltitle{Hierarchical Expert Networks for Meta-Learning}



\icmlsetsymbol{equal}{*}

\begin{icmlauthorlist}
\icmlauthor{Heinke Hihn}{uulm}
\icmlauthor{Daniel A.~ Braun}{uulm}

\end{icmlauthorlist}

\icmlaffiliation{uulm}{Institute for Neural Information Processing, Ulm University, Ulm, Germany}

\icmlcorrespondingauthor{Heinke Hihn}{heinke.hihn@uni-ulm.de}
\icmlkeywords{Meta-Learning, Regression, Classification, Reinforcement Learning, Information Theory}

\vskip 0.3in
]



\printAffiliationsAndNotice{}  

\begin{abstract}
The goal of meta-learning is to train a model on a variety of learning tasks, such that it can adapt to new problems within only a few iterations. Here we propose a principled information-theoretic model that optimally partitions the underlying problem space such that specialized expert decision-makers solve the resulting sub-problems. To drive this specialization we impose the same kind of information processing constraints both on the partitioning and the expert decision-makers. We argue that this specialization leads to efficient adaptation to new tasks. To demonstrate the generality of our approach we evaluate three meta-learning domains: image classification, regression, and reinforcement learning.
\end{abstract}

\section{Introduction}
Recent machine learning research has shown impressive results on incredibly diverse tasks from problem classes such as pattern recognition, reinforcement learning, and generative model learning \cite{devlin2018bert,Mnih2015,Schmidhuber2015}. These success stories typically have two computational luxuries in common: a large database with thousands or even millions of training samples and a long and extensive training period. Applying these pre-trained models to new tasks na\"{i}vely usually leads to poor performance, as with each new incoming batch of data, expensive and slow re-learning. In contrast to this, humans can learn from few examples and excel at adapting quickly \cite{jankowski2011meta}, for example in motor tasks \cite{braun2009motor} or at learning new visual concepts \cite{lake2015human}. 

Sample-efficient adaptation to new tasks can is a form of meta-learning or ``learning to learn'' \cite{thrun2012learning, schmidhuber1997shifting, caruana1997multitask} and is an ongoing and active field of research--see e.g.,~ \cite{koch2015siamese, vinyals2016matching, Finn2017model, ravi2017optimization, ortega2019meta, botvinick2019reinforcement, yao2019hierarchically}. Meta-learning has multiple definitions, but a common point is that the system learns on two levels, each with different time scales: slow meta-learning across different tasks, and fast learning to adapt to each task individually.

\begin{figure*}[t!]
\begin{center}
\includegraphics[width=.2\textwidth]{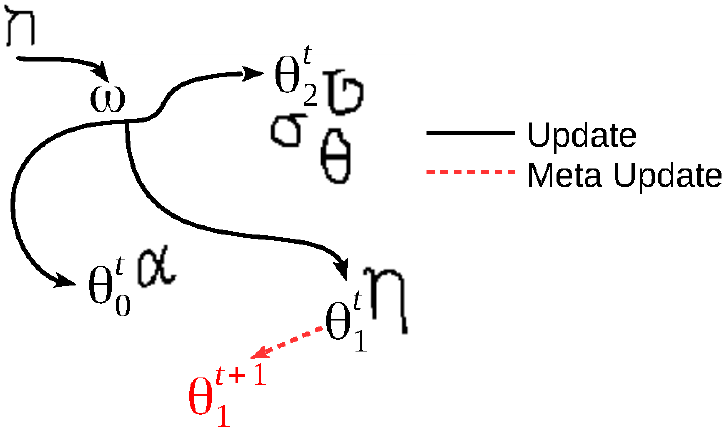}
\hspace{1em}
\includegraphics[width=.7\textwidth, trim={2cm 24.5cm 0.75cm 0.75cm}, clip]{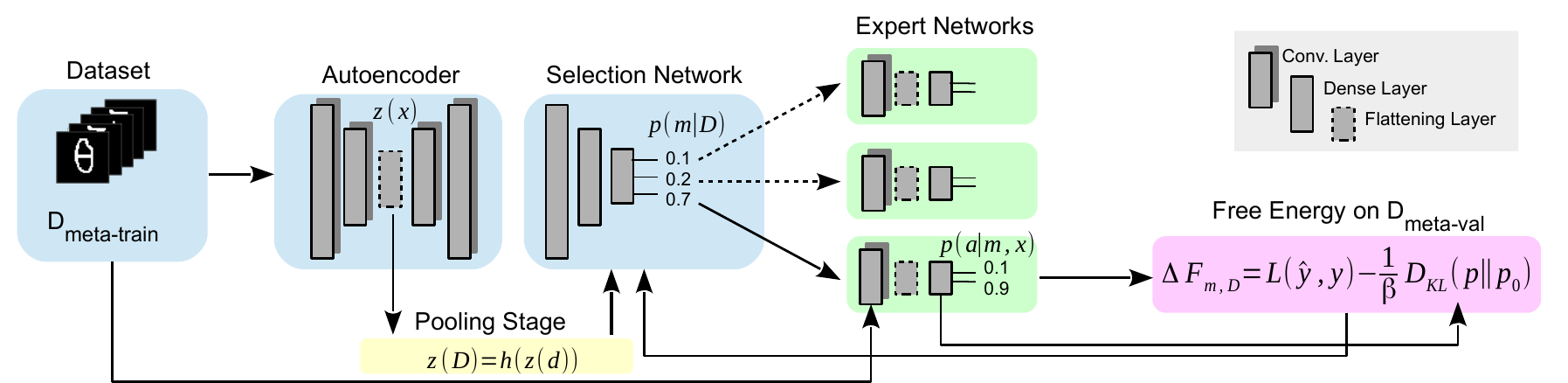}
\end{center}
\caption{\textbf{Left}: The selector assigns the new input encoding to one of the three experts $\theta_0$, $\theta_1$, or $\theta_2$, depending on the similarity of the input to previous inputs seen by the experts. \textbf{Right:} Our proposed method consists of three main stages. First, we feed the training dataset $D_\text{train}$ through a convolutional autoencoder to find a latent representation $z(d_i)$ for each $d_i \in D_\text{train}$, which we get by flattening the preceding convolutional layer (``flattening layer"). We apply a pooling function to the resulting set of image embeddings which serves as input to the selection network.} 
\label{fig:model}
\end{figure*}

Here, we propose a novel information-theoretic learning paradigm for hierarchical meta-learning systems. The method we propose comes from a single unified framework and can be readily applied to supervised meta-learning and to meta-reinforcement learning. Our method finds an optimal soft partitioning of the problem space by imposing information-theoretic constraints on both the process of expert selection and on the expert specialization. We argue that these constraints drive an efficient division of labor in systems that have limited information processing power, where we make use of information-theoretic bounded rationality \cite{Ortega2013}. When the model is presented with previously unseen tasks it assigns them to experts specialized on similar tasks -- see Figure \ref{fig:model}. Additionally, expert networks specializing in only a subset of the problem space allow for smaller neural network architectures with only a few units per layer. To split the problem space and to assign the partitions to experts, we learn to represent tasks through a common latent embedding, that is then used by a selector network to distribute the tasks to the experts.

The outline of this paper is as follows: first, we introduce bounded rationality and meta-learning, next we introduce our novel approach and derive applications to classification, regression, and reinforcement learning. We evaluate our method against current meta-learning algorithms and perform additional ablation studies. Finally, we conclude.

\section{Information-Processing Constraints in Hierarchical Learning Systems}
An important concept in decision making is the notion of utility \cite{VonNeumann2007}, where an agent picks an action $a_x^* \in \mathcal{A}$ such that it maximizes their utility in some context $s \in \mathcal{S}$, i.e.,~ $a^*_x = \argmax_a \mathbf{U}(x, a)$,
where the utility is a function $\mathbf{U}(x, a)$ and the states distribution $p(s)$ is known and fixed. Trying to solve this optimization problem na\"{i}vely leads to an exhaustive search over all possible $(a,x)$ pairs, which is in general a prohibitive strategy. Instead of finding an optimal strategy, a \textit{bounded-rational decision-maker} optimally trades off expected utility and the processing costs required to adapt. In this study we consider the information-theoretic free-energy principle \cite{Ortega2013} of bounded rationality, where an upper bound on the Kullback-Leibler divergence $\DKL(p(a|x)||p(a)) = \sum_{a}{p(a|x) \log{\frac{p(a|x)}{p(a)}}}$ between the agent's prior distribution $p(a)$ and the posterior policy $p(a|x)$ model the decision-maker's resources, resulting in the following constrained optimization problem:
\begin{align}
\max_{p(a|x)} \sum_{x,a} p(x){p(a|x)\mathbf{U}(x, a)} \\ \text{ s.t. } \mathbb{E}_{p(x)}\left[\DKL(p(a|x)||p(a))\right] \leq \text{B}.
\end{align}
This constraint defines a regularization on $p(a|x)$. We can transform this into an unconstrained variational problem by introducing a Lagrange multiplier  $\beta \in \mathbb{R}^+$:
\begin{equation}
\max_{p(a|x} \mathbb{E}_{p(x|a)}\left[\mathbf{U}(x, a)\right] - \frac{1}{\beta}\mathbb{E}_{p(x)}\left[\DKL(p(a|x)||p(a))\right].
\label{eq:br_obj}
\end{equation}
For $\beta \rightarrow \infty$ we recover the maximum utility solution and for $\beta \rightarrow 0$ the agent can only act according to the prior. The optimal prior in this case is the marginal $p(a) = \sum_{x \in X}{p(x) p(a|x)}$ \cite{Ortega2013}. 

Aggregating bounded-rational agents by a selection policy allows for solving optimization problems that exceed the capabilities of the individual decision-makers \cite{Genewein2015}. To achieve this, the search space is split into partitions such that each partition can be solved by a decision-maker. A two-stage mechanism is introduced: The first stage is an expert selection policy $p(m|x)$ that chooses an expert $m$ given a state $x$ and the second stage chooses an action according to the expert's posterior policy $p(a|x,m)$. The optimization problem given by Equation \eqref{eq:br_obj} can be extended to incorporate a trade-off between computational costs and utility in both stages:   
\begin{equation}
\label{eq:par_mutual}
\max_{p(a|x,m), p(m|x)} \E[\mathbf{U}(x,a)] - \frac{1}{\beta_1}I(X;M) - \frac{1}{\beta_2}I(A;X|M)
\end{equation}
where $\beta_1$ is the resource parameter for the expert selection stage and $\beta_2$ for the experts. $I(\cdot;\cdot)$ is the mutual information between the two random variables. To measure a decision-maker's performance on terms of the utility vs.~ cost trade off we use the free-energy difference defined as 
\begin{equation} 
\F = \E_{p}[\mathbf{U}] - \frac{1}{\beta}\DKL(p\vert\vert q),
\end{equation}
where $p$ is the posterior and $q$ the decision-maker's prior policy. The marginal distribution $p(a|x)$ defines a \emph{mixture-of-experts} \cite{Jordan1994, Jacobs1991} policy given by the posterior distributions $p(a|s,m)$ weighted by the responsibilities determined by the Bayesian posterior $p(x|m)$. Note that $p(x|m)$ is not determined by a given likelihood model, but is the result of the optimization process~\eqref{eq:par_mutual}.

\section{Meta Learning}
\label{sec:meta}
We can divide Meta-learning algorithms roughly into Metric-Learning  \cite{koch2015siamese, vinyals2016matching,snell2017prototypical}, Optimizer Learning \cite{ravi2017optimization,Finn2017model,zintgraf2018caml, rothfuss2018promp}, and Task Decomposition Models \cite{lan2019meta, vezhnevets2019options}. Our approach depicted in Figure~\ref{fig:model} can be seen as a member of the latter group.
 
\subsection{Meta Supervised Learning}
In a supervised learning task we are usually interested in a dataset consisting of multiple input and output pairs $D = \{(x_i, y_i)\}^N_{i=1}$ and the learner's task is to find a function $f(x)$ that maps from input to output, for example through a deep neural network. To do this, we split the dataset into training and test sets and fit a set of parameters $\theta$ on the training data and evaluate on test data using the learned function $f_\theta(x)$. In meta-learning, we are instead working with meta-datasets $\mathcal{D}$, each containing regular datasets split into training and test sets. We thus have different sets for meta-training,  meta-validation, and meta-test, i.e.,~ $\mathcal{D} = \{D_\text{train}, D_\text{val},D_\text{test}\}$. The goal is to train a learning procedure (the meta-learner) that can take as input one of its training sets $D_\text{train}$ and produce a classifier (the learner) that achieves low prediction error on its corresponding test set $D_\text{test}$. The meta-learning is then updated using performance measure based the learner's performance on $D_{\text{val}}$. This may not always be the case, but our work (among others, e.g.,~ \cite{Finn2017model}) follow this paradigm. The rationale being that the meta-learner is trained such that it implicitly optimizes the base learner's generalization capabilities. 


\subsection{Meta Reinforcement Learning}
First, we give a short introduction to reinforcement learning in general, and then we show how this definition can be extended to meta reinforcement learning problems. 
We model sequential decision problems by defining a Markov Decision Process as a tuple $(\mathcal{S}, \mathcal{A}, P, r)$, where $\mathcal{S}$ is the set of states, $\mathcal{A}$ the set of actions, $P: \mathcal{S} \times \mathcal{A} \times \mathcal{S} \rightarrow [0,1]$ is the transition probability, and $r: \mathcal{S} \times \mathcal{A} \rightarrow \mathbb{R}$ is a reward function. The aim is to find the parameter $\theta$ of a policy $\pi_\theta$ that maximizes the expected reward:

\begin{equation}
\theta^* = \argmax_{\theta} \underbrace{\mathbb{E}_{\pi_\theta}\left[\sum_{t=0}^\infty r(s_t, a_t)\right]}_{J(\pi_\theta)}.
\end{equation}
We define $r(\tau) = \sum_{t=0}^\infty r(s_t, a_t)$ as the cumulative reward of trajectory $\tau = \{(s_t, a_t)\}_{i=0}^\infty$,  which is sampled by acting according to the policy $\pi$, i.e.,~ $(s, a)\sim \pi(\cdot|s),$ and $s_{t+1} \sim P(\cdot|s_t, a_t)$.
We model learning in this environment as reinforcement learning \cite{Sutton2018}, where an agent interacts with an environment over some (discrete) time steps $t$. At each time step $t$, the agent finds itself in a state $s_t$ and selects an action $a_t$ according to the policy $\pi(a_t|s_t)$. In return, the environment transitions to the next state $s_{t+1}$ and generates a scalar reward $r_t$. This process continues until the agent reaches a terminal state after which the process restarts. The goal of the agent is to maximize the expected return from each state $s_t$, which is typically defined as the infinite horizon discounted sum of the rewards.

In meta reinforcement learning the problem is given by a set of tasks $t_i \in T$, where MDP $t_i = (\mathcal{S}, \mathcal{A}, P_i, r_i)$  define tasks $t_i$   as described earlier. We are now interested in finding a set of policies $\Theta$ that maximizes the average cumulative reward across all tasks in $T$ and generalizes well to new tasks sampled from a different set of tasks $T'$.

\section{Expert Networks for Meta-Learning}
\begin{algorithm}[t]
\caption{Expert Networks for Supervised Meta-Learning.}
\label{alg:meta}
\begin{algorithmic}[1]
\STATE \textbf{Input}: Data Distribution $p(\mathcal{D})$, number of samples $K$, batch-size $M$, training episodes $N$\;
\STATE \textbf{Hyper-parameters}: resource parameters $\beta_1$, $\beta_2$, learning rates $\eta_{x}$, $\eta_x$ for selector and experts\;
\STATE Initialize parameters $\theta, \vartheta$\;
\FOR{$i$ = 0, 1, 2, ..., $N$}
\STATE Sample batch of $M$ datasets $D_i \thicksim p(\mathcal{D})$, each consisting of a training dataset $D_\text{{meta-train}}$ and a meta-validation dataset\; $D_\text{{meta-val}}$ with $2K$ samples each \;
\FOR{$D \in D_i$}
\STATE Find Latent Embedding $z(D_\text{{meta-train}})$ \;
\STATE Select expert $m \thicksim p_{\theta}(m\vert z(D_\text{{meta-train}})$ \;
\STATE Compute $\hat{f}(m, D_\text{val})$ \;
\ENDFOR\
\STATE Update selection parameters $\theta$ with $\hat{f}(m, D_\text{val})$ \;
\STATE Update Autoencoder with positive samples in $D_i$ \;
\STATE Update experts $m$ with assigned $D_\text{{meta-train}}$\;
\ENDFOR\
\STATE \textbf{return} $\theta$, $\vartheta$\;
\end{algorithmic}
\end{algorithm}

Information-theoretic bounded rationality postulates that hierarchies and abstractions emerge when agents have only limited access to computational resources \cite{Genewein2015}, e.g.,~ limited sampling complexity \cite{Hihn2018} or limited representational power \cite{hihn2019}. We will show that forming such abstractions equips an agent with the ability of learning the underlying problem structure and thus enables learning of unseen but similar concepts. The method we propose comes out of a unified optimization principle and has the following important features: 
\begin{enumerate}
\item A regularization mechanism to enforce the emergence of expert policies.
\item A task compression mechanism to extract relevant task information.
\item A selection mechanism to find the most efficient expert for a given task.
\item A regularization mechanism to improve generalization capabilities.
\end{enumerate}

\subsection{Latent Task Embeddings}
Note that the selector assigns a complete dataset to an expert and that this can be seen as a meta-learning task, as described in \cite{ravi2017optimization}. To do so, we must find a feature vector $z(d)$ of the dataset $d$. This feature vector must fulfill the following desiderata: 1) invariance against permutation of data points in $d$, 2) high representational capacity, 3) efficient computability, and 4) constant dimensionality regardless of sample size $K$. In the following we propose such features for image classification, regression, and reinforcement learning problems.

For image classification we propose to pass the positive images in the dataset through a convolutional autoencoder and use the outputs of the bottleneck layer. Convolutional Autoencoders are generative models that learn to reconstruct their inputs by minimizing the Mean-Squared-Error between the input and the reconstructed image (see e.g.,~ \cite{chen2019closer}). In this way we get similar embeddings $z(d)$ for similar inputs belonging to the same class. The latent representation is computed for each positive sample in $d$ and then passed through a pooling function $h(z(d))$ to find a single embedding for the complete dataset--see figure \ref{fig:model} for an overview of our proposed model. While in principle functions such as mean, max, and min can be used, we found that max-pooling yields the best results. The authors of \cite{yao2019hierarchically} propose a similar feature set.

For regression we define a similar feature vector. We transform the $K$ training data points into a feature vector $z(d)$ by binning the points into $N$ bins according to their $x$ value and collecting the $y$ value. If more than one point falls into the same bin, we average the $y$ values, thus providing invariance against the order of the data points in $D_{\text{train}}$. We use this feature vector to assign each data set to an expert according to $p_\theta(x|z(d))$. 

In the reinforcement learning setting we use a dynamic recurrent neural network (RNN) with LSTM units \cite{hochreiter1997long} to classify trajectories. We feed the RNN with $(s_t, a_t, r_t, t)$ tuples to describe the underlying Markov Decision Process describing the task. At $t = 0$ we sample the expert $x$ according to the learned prior distribution $p(x)$, as there is no information available so far. The authors of \cite{lan2019meta} propose a similar feature set. 

\subsection{Specialization in Supervised Learning}
Combining multiple experts can often be beneficial \cite{Kuncheva2004}, e.g. in Mixture-of-Experts \cite{Yuksel2012} or Multiple Classifier Systems \cite{Bellmann2018}. Our method can be interpreted as a member of this family of algorithms. 

We define the utility as the negative prediction loss, i.e. $U(f_x(d), y) = -\mathcal{L}(f_x(d),y)$, where $f_x(d)$ is the prediction of the expert $x$ given the input data point $d$ (in the following we will use the shorthand $\hat{y}_x$) and $y$ is the ground truth. We define the cross-entropy loss $\mathcal{L}(\hat{y}_x, y,)= -\sum_i y_i \log \hat{y_i}_x$ as a  performance measure for classification and the mean squared error $\mathcal{L}(\hat{y}_x, y) = \sum_i (\hat{y}_i-y_i)^2 $ for regression. The objective for expert selection thus is given by

\begin{equation}
\max_\theta \E_{p_\theta(x|d)}\left[\hat{f} - \frac{1}{\beta_1}\log\frac{p_\theta(x|d)}{p(x)}\right],
\end{equation}
where $\hat{f} = \mathbb{E}_{p_\vartheta(\hat{y}_x|x,s)}\left[-\mathcal{L}(\hat{y}_x,y) - \frac{1}{\beta_2}\log\frac{p_\vartheta(\hat{y}_x|s,x)}{p(\hat{y}_x|x)}\right]$, i.e. the free energy of the expert and $\theta, \vartheta$ are the parameters of the selection policy and the expert policies, respectively. Analogously, the action selection objective for each expert $x$ is defined by
\begin{equation}
\max_\vartheta \E_{p_\vartheta(\hat{y}_x|x,s)}\left[-\mathcal{L}(\hat{y}_x,y) - \frac{1}{\beta_2}\log\frac{p_\vartheta(\hat{y}_x|s,x)}{p_(\hat{y}_x|x)}\right].
\end{equation}
This regularization technique can be seen as a form of output regularization \cite{pereyra2017regularizing}.
\subsection{Specialization in Reinforcement Learning}

\begin{algorithm}[t]
\caption{Expert Networks for Meta-Reinforcement Learning.}
\label{alg:metarl}
\begin{algorithmic}[1]
\STATE \textbf{Input}: Environment Distributions $p(T)$ and $p(T')$, number of roll-outs $K$, batch-size $M$, training episodes $N$, number of tuples $L$ used for expert selection\;
\STATE \textbf{Hyper-parameters}: resource parameters $\beta_1$, $\beta_2$, learning rates $\eta_{x}$, $\eta_x$ for selector and experts\;
\STATE Initialize parameters $\theta, \vartheta$\;
\FOR{$i$ = 1, 2, 3, ..., $N$}
\STATE Sample batch of $M$ environments $E_i^{\text{train}} \thicksim p(T)$ and $E_i^{\text{val}} \thicksim p(T')$\;
\FOR{$E \in E_i^{\text{train}}$}
\FOR{$k$ = 1, 2, 3, ..., $K$}
\STATE Collect $\tau = \{(x_t, a_t, r_t, t)\}_{t=1}^L$ tuples by following random expert \;
\STATE Select expert $m \thicksim p_{\theta}(m\vert\tau)$ with RNN policy \;
\STATE Collect trajectory $\tau_k = \{(x_t, a_t, r_t, t)\}_{t=L}^T$ by following $p_\vartheta(a|x,m)$
\ENDFOR\
\STATE Compute $F_t = \sum_{l=0}^T\gamma^l f(x_{t+l}, m_{t+l}, a_{t+l})$ for trajectories $\tau$\;
\STATE where $f(x,m,a) = r_{\text{train}}(x, a) - \frac{1}{\beta_2} \log\frac{p_{\vartheta}(a\vert x,m)}{p(a\vert m)}$.
\ENDFOR\
\STATE Compute $\bar{F}_t = \sum_{l=0}^T\gamma^l \bar{f}(x_{t+l}, m_{t+l})$ with \;
\STATE {\small $\bar{f}(x,m) = \mathbb{E}_{p_\vartheta(a\vert x,m)}\left[r_{\text{val}}(x,a) - \frac{1}{\beta_2} \log \frac{p(a\vert x, m)}{p(a\vert m)}\right]$ }\;
\STATE Update selection parameters $\theta$ with $\bar{F}$ collected in batch $i$ \;
\STATE Update experts $m$ with roll-outs collected in batch $i$\;
\ENDFOR\
\STATE \textbf{return} $\theta$, $\vartheta$\;
\end{algorithmic}
\end{algorithm} 
In the following we will derive our algorithm for specialization in hierarchical reinforcement learning agents. Note that in the reinforcement learning setup the reward function $r(x,a)$ defines the utility $\mathbf{U}(x,a)$. In maximum entropy RL (see e.g.,\, Haarnoja et al.\, (2017) \cite{haarnoja2017reinforcement}) the regularization penalizes deviation from a fixed uniformly distributed prior, but in a more general setting we can discourage deviation from an arbitrary prior policy by optimizing for: 
\begin{equation}
\label{eq:maxentrl}
\max_{p}\mathbb{E}_{p}\left[\sum_{t=0}^\infty\gamma^t \left( r(x_t, a_t) -  \frac{1}{\beta} \log \frac{p(a_t\vert x_t)}{p(a)}\right)\right],
\end{equation}
where $\beta$ trades off between reward and entropy, such that $\beta \rightarrow \infty$ recovers the standard RL value function and $\beta \rightarrow 0$ recovers the value function under a random policy. 

To optimize the objective \eqref{eq:maxentrl} we define two separate kinds of value function, $V_\phi$ for the selector and one value function $V_\varphi$ for each expert. Thus, each expert is an actor-critic with separate actor and critic networks. Similarly, the selector has an actor-critic architecture, where the actor network selects experts and the critic learns to predict the expected free energy of the experts depending on a state variable. The selector's policy is represented by $p_\theta$, while each expert's policy is represented by a distribution $p_\vartheta$.

%
%

In standard reinforcement learning a common technique to update a parametric policy representation $p_\omega(a|x)$ with parameters $\omega$, is to use policy gradients that optimize the  cumulative reward
$
J(\omega) = \mathbb{E}\left[ p_\omega(a|x) V_\psi(x)\right]
$
expected under the critic's prediction $V_\psi(x)$, by following the gradient
$
\nabla_\omega J(\omega) = \mathbb{E}\left[\nabla_\omega \log p_\omega(a|x) V_\psi(x) \right] .
$
This policy gradient formulation \cite{Sutton2000} is prone to producing high variance gradients. A common technique to reduce the variance is to formulate the updates using the advantage function instead of the reward \cite{arulkumaran2017deep}. The advantage function $A(a, x)$ is a measure of how well a certain action $a$ performs in a state $x$ compared to the average performance in that state, i.e.,~$A(a, x) = Q(x, a) - V_\psi(x)$. Here, $V(x)$ is the value function and is a measure of how well the agent performs in state $x$, and $Q(x, a)$ is an estimate of the cumulative reward achieved in state $x$ when the agent executes action $a$. Instead of learning the value and the Q function, we can approximate the advantage function solely based on the critic's estimate $V_\psi(x)$ by noting that $Q(x_t, a_t) \approx r(x_t, a_t) + \gamma V_{\psi}(x_{t+1})$.
Similar to the standard policy update based on the advantage function,
the expert selection stage can be formulated by optimizing the expected advantage $\mathbb{E}_{p_\vartheta(a|x,m)} \left[A_m(x,a)\right]$ for expert $m$ with
\begin{equation}
A_m(x_t,a_t) = f(x_t,m,a_t) + \gamma V_\varphi(x_{t+1})-V_\varphi(x_t).
\end{equation}
Accordingly, we can define an expected advantage function $\mathbb{E}\left[p_\theta(m|x) \bar{A}(x,m)\right]$ for the selector with
\begin{equation}
\bar{A}(x,m) = \mathbb{E}_{p_\vartheta(a|x,m)} \left[ A_m(x,a) \right] ,
\end{equation}
where $m_t$ denotes the expert $m$ selected at time $t$. We estimate the double expectation by Monte Carlo sampling, where in practice we use a single $(x, m, a)$ tuple for $\hat{f}(x,m)$, which enables us to employ our algorithm in an on-line optimization fashion.
\begin{table}[t!]
\centering
\resizebox{0.49\textwidth}{!}{
\begin{tabular}{*6c}
\toprule
\multicolumn{6}{c}{\textsc{Omniglot Few-Shot}}  \\
\toprule
{} & \multicolumn{3}{c}{\textsc{One Conv. Block Baselines}} &  \multicolumn{2}{c}{\textsc{Methods}}  \\
\midrule
{} & \textsc{Pre-Training} & \textsc{MAML} &  \textsc{Matching Nets} & \textsc{MAML} & \textsc{Matching Nets}  \\
\midrule
\textbf{K} & \textbf{\% Acc} &  \textbf{\% Acc} &  \textbf{\% Acc} &\textbf{\% Acc} & \textbf{\% Acc} \\
\midrule
\textbf{1} & 50.6 ($\pm$ 0.03) & 81.2 ($\pm$ 0.03) &  52.7 ($\pm$ 0.05) & 95.2 ($\pm$ 0.03) &  95.0 ($\pm$ 0.01)\\
\textbf{5} & 54.1 ($\pm$ 0.09) & \textbf{88.0} ($\pm$ 0.01) & 55.3 ($\pm$ 0.04) & 99.0 ($\pm$ 0.01) &  98.7 ($\pm$ 0.01)\\
\textbf{10} & 55.8 ($\pm$ 0.02) & 89.2 ($\pm$ 0.01) & 60.9 ($\pm$ 0.06) & 99.2 ($\pm$ 0.01) &  99.4 ($\pm$ 0.01) \\
\bottomrule
\multicolumn{5}{c}{\textsc{Our Method}} \\
\midrule
\multicolumn{5}{c}{\textsc{Number of Experts}} \\
\midrule
{} &  \multicolumn{2}{c}{\textbf{2}} & \multicolumn{2}{c}{\textbf{4}}\\
\midrule
{} & \textbf{\% Acc} & \textbf{I(M;X)} & \textbf{\% Acc} & \textbf{I(M;X)} \\
\midrule
\textbf{1} & 66.4 ($\pm$ 0.02) & 0.99 ($\pm$ 0.01) & 75.8 ($\pm$ 0.02) & 1.96 ($\pm$ 0.01)\\
\textbf{5} & 67.3 ($\pm$ 0.01) & 0.93 ($\pm$ 0.01) & 75.5 ($\pm$ 0.01) & 1.95 ($\pm$ 0.10) \\
\textbf{10} & 76.2 ($\pm$ 0.04)  & 0.95 ($\pm$ 0.30) & 86.7 ($\pm$ 0.01) & 1.90 ($\pm$ 0.03) \\
\midrule
{} & \multicolumn{2}{c}{\textbf{8}} & \multicolumn{2}{c}{\textbf{16}} \\
\midrule
{} & \textbf{\% Acc} & \textbf{I(M;X)} & \textbf{\% Acc} & \textbf{I(M;X)} \\
\midrule
\textbf{1} & 77.3 ($\pm$ 0.01) & 2.5 ($\pm$ 0.02) & \textbf{82.8} ($\pm$ 0.01) & 3.2 ($\pm$ 0.03) \\
\textbf{5} & 78.4 ($\pm$ 0.01) & 2.7 ($\pm$ 0.01) & 85.2 ($\pm$ 0.01) & 3.3 ($\pm$ 0.02) \\
\textbf{10} & 90.1 ($\pm$ 0.01) & 2.8 ($\pm$ 0.02) & \textbf{95.9} ($\pm$ 0.01) & 3.1 ($\pm$ 0.02)\\
\bottomrule
\end{tabular}}
\caption{Classification accuracy after 10 gradient steps on the validation data. Adding experts consistently improves performance, obtaining the best results with an ensemble of 16 experts. Pre-training refers to a single expert system trained on the complete dataset. Our method outperforms the pre-training, Matching Nets, and the MAML baseline (see Section \ref{sec:regressionresults} for experimental details), when the network architecture is reduced to a single convolution block. This corresponds to our expert network architecture. Using the suggested architectures by the respective studies, we achieve classification accuracy $\geq 95\%$.}
\label{tab:classtable}
\end{table}

In the RL setup the reward function $r(x,a)$ defines the utility $\mathbf{U}(x,a)$. To optimize the objective we define value functions, $V_\phi$ for the selector and one value function $V_\varphi$ for each expert. Thus, the selector and each expert is an actor-critic architecture. We define a discounted free energy as 
\begin{equation}
F_t = \sum_{l=0}^T\gamma^l \left(r(x_{t+l}, a_{t+l}) - \frac{1}{\beta_2} \log\frac{p_{\vartheta}(a_{t+l}\vert x_{t+l},m_{t+l})}{p(a_{t+l}\vert m_{t+l})}\right),\end{equation} 
which we learn through a value function $V_\varphi$ for each expert. We define the selector's discounted free energy \begin{equation}\bar{F}_t = \sum_{l=0}^T\gamma^l \bar{f}(x_{t+l}, m_{t+l}),\end{equation} with $\bar{f}(x,m) = \mathbb{E}_{p_\vartheta(a\vert x,m)}[r(x,a) - \frac{1}{\beta_2} \log \frac{p_\vartheta(a\vert x, m)}{p(a\vert m)}]$ that is learned through the selector's value function $V_\phi$.
Now let $p_\theta(m|x)$ be the selection policy and $p_\vartheta(a|x,m)$ the expert policy. The expert selection stage is optimizing the expected advantage $\mathbb{E}\left[p_\vartheta(a|x,m) A_m(x,a)\right]$ for expert $m$ with \begin{equation}A_m(x_t,a_t) = f(x_t,m,a_t) + \gamma V_\varphi(x_{t+1})-V_\varphi(x_t).\end{equation} Accordingly, we define an expected advantage function $\mathbb{E}\left[p_\theta(m|x) \bar{A}(x,m)\right]$ for the selector with \begin{equation}\bar{A}(x,m) = \mathbb{E}_{p_\vartheta(a|x,m)} \left[ A_m(x,a) \right].\end{equation} We find the policy parameters by gradient descent on the objectives.
\begin{figure}[t!]
\centering
\begin{minipage}{0.225\textwidth}
\centering
\includegraphics[width=\linewidth]{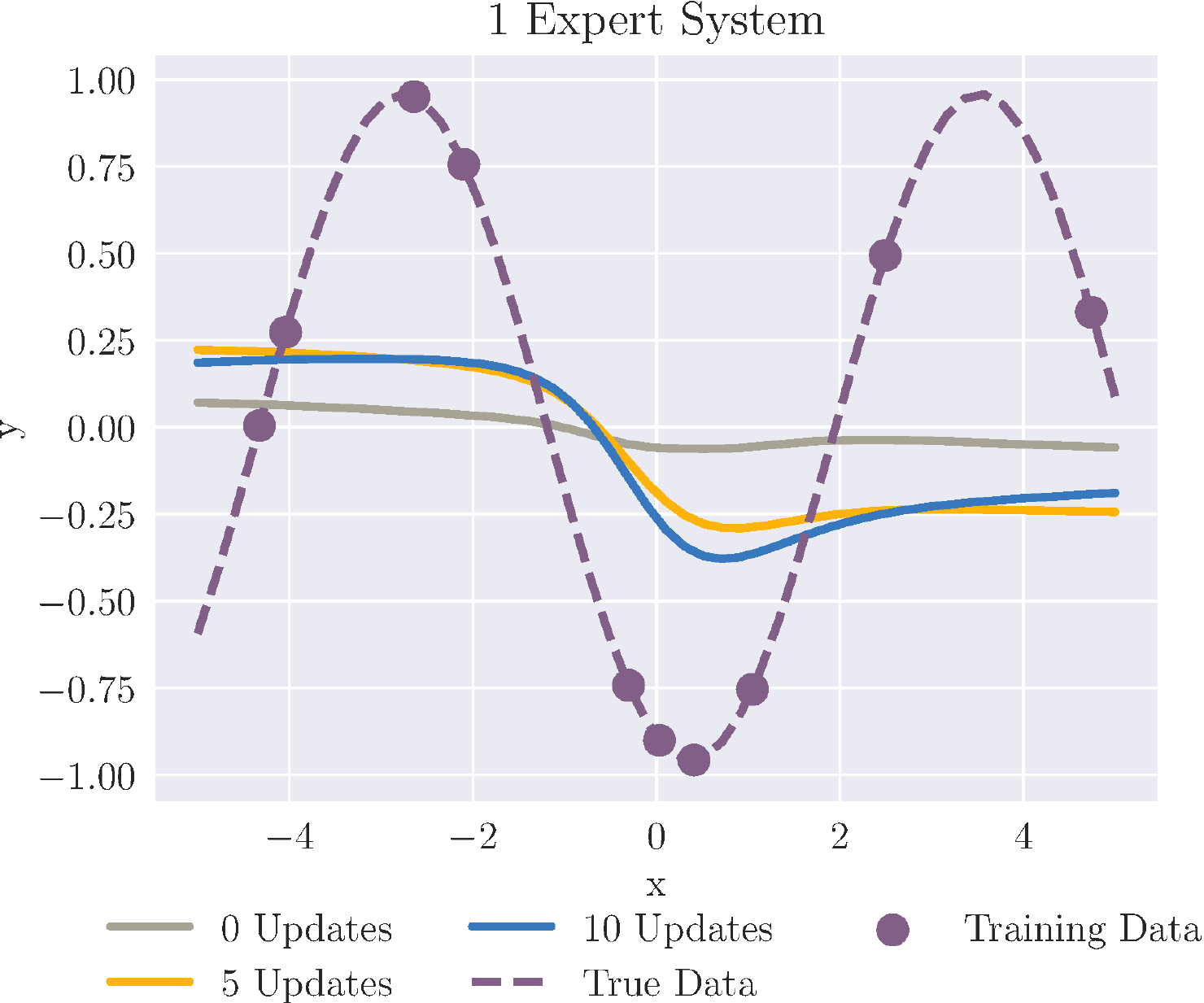}
\end{minipage}
\begin{minipage}{0.225\textwidth}
\centering
\includegraphics[width=\linewidth]{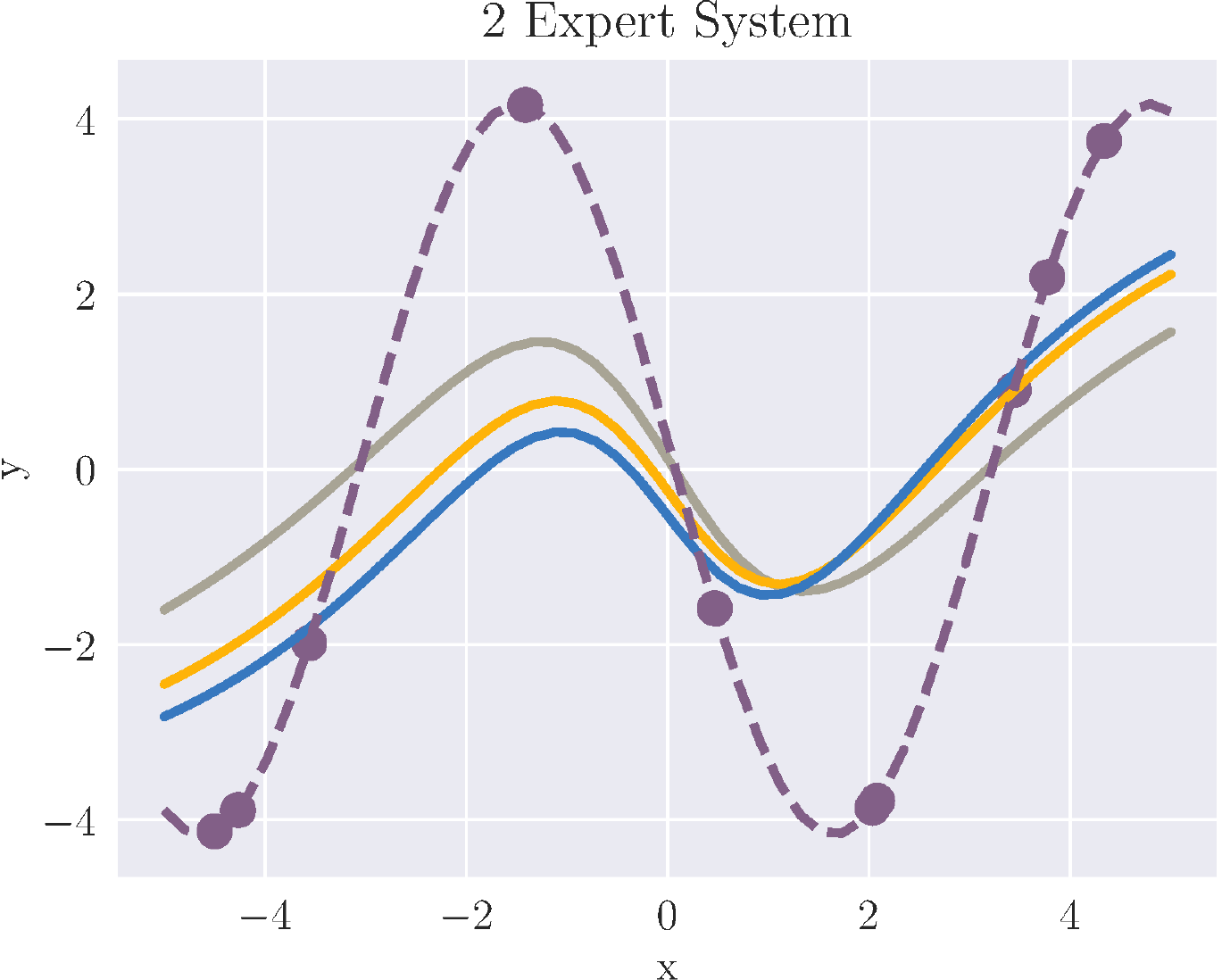}
\end{minipage}
\begin{minipage}{0.225\textwidth}
\centering
\includegraphics[width=\linewidth]{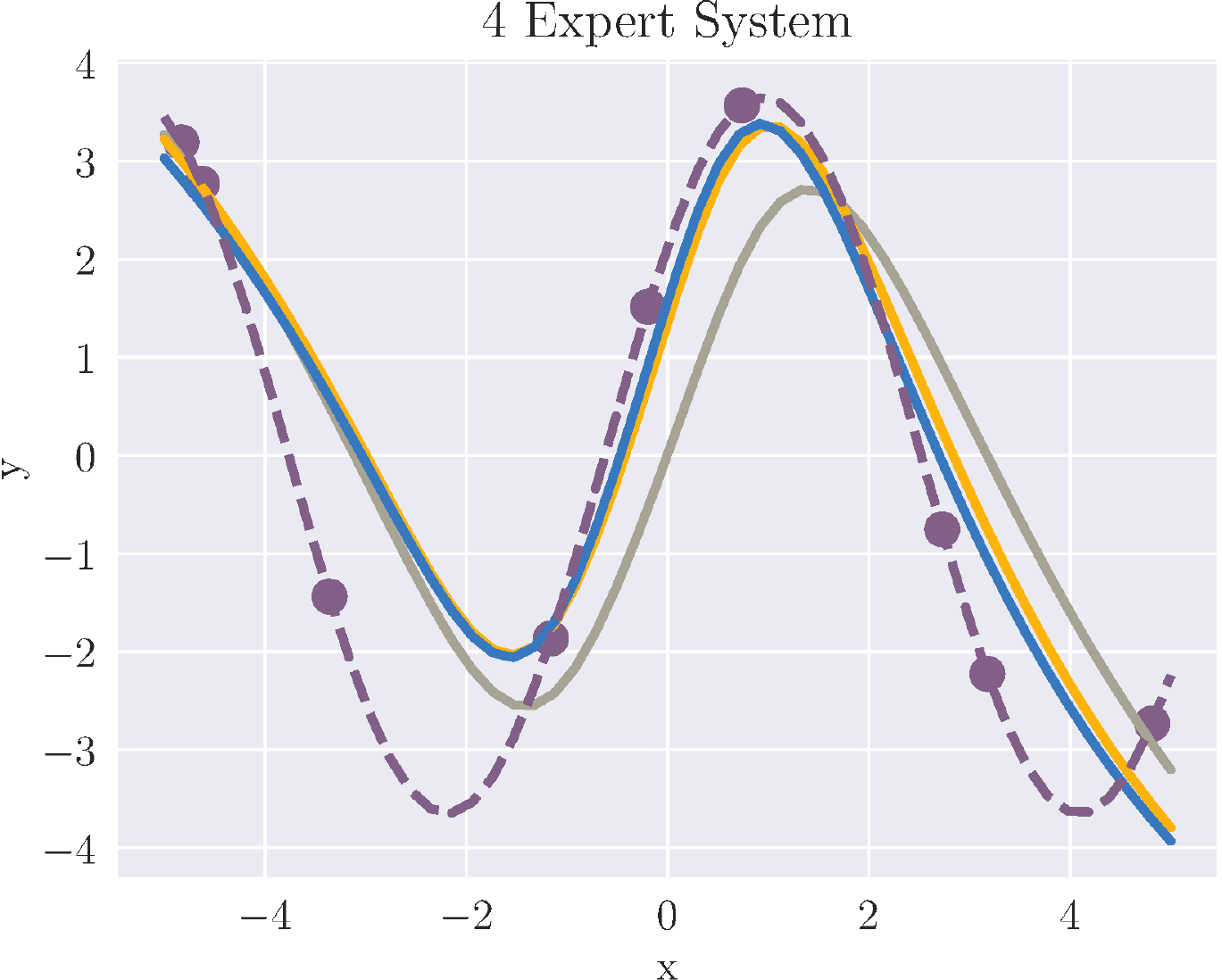}
\end{minipage}
\begin{minipage}{0.225\textwidth}
\centering
\includegraphics[width=\linewidth]{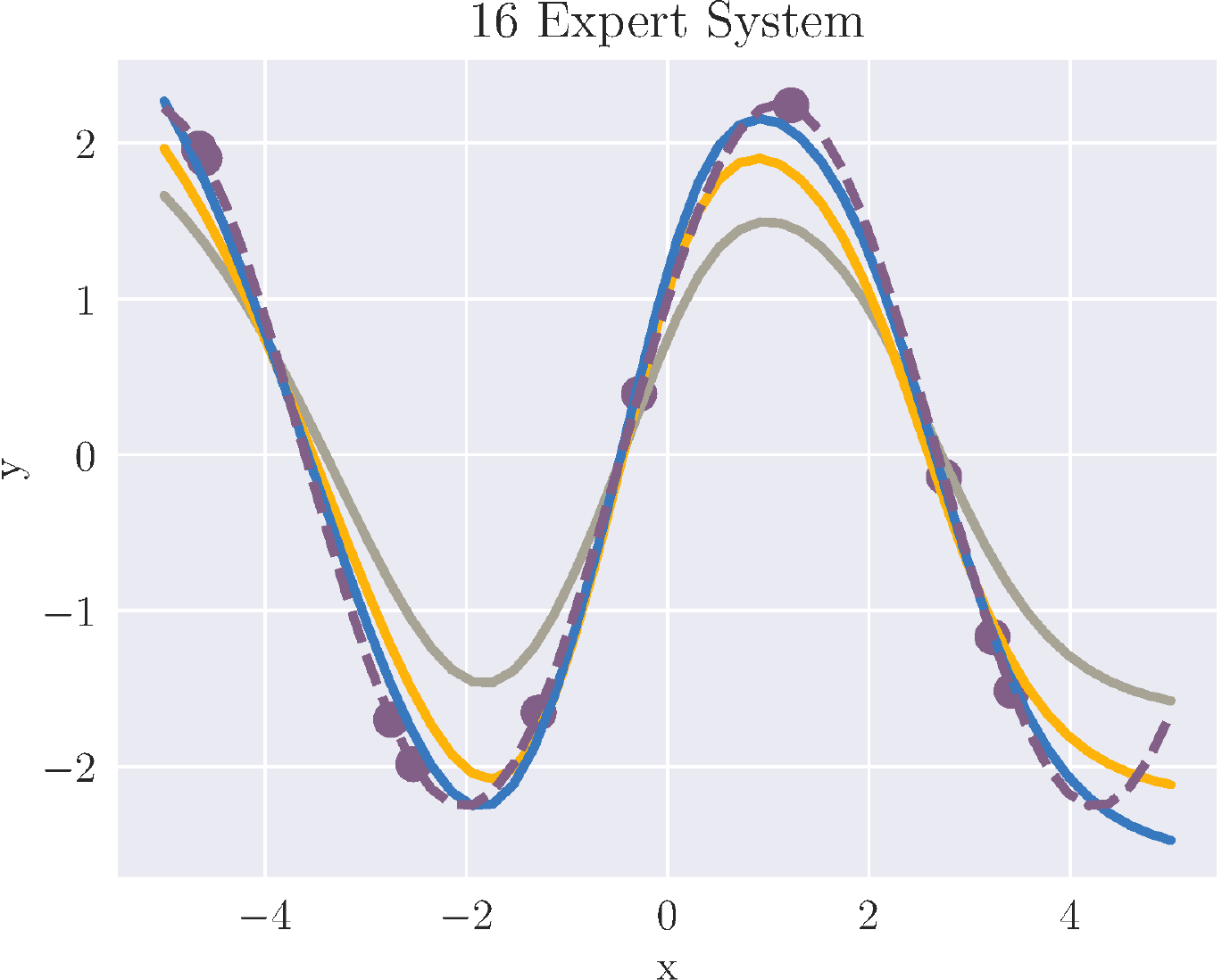}
\end{minipage}
\begin{minipage}{\linewidth}
\centering\
\includegraphics[width=.99\linewidth, trim={0cm 0cm 0cm 0cm}, clip]{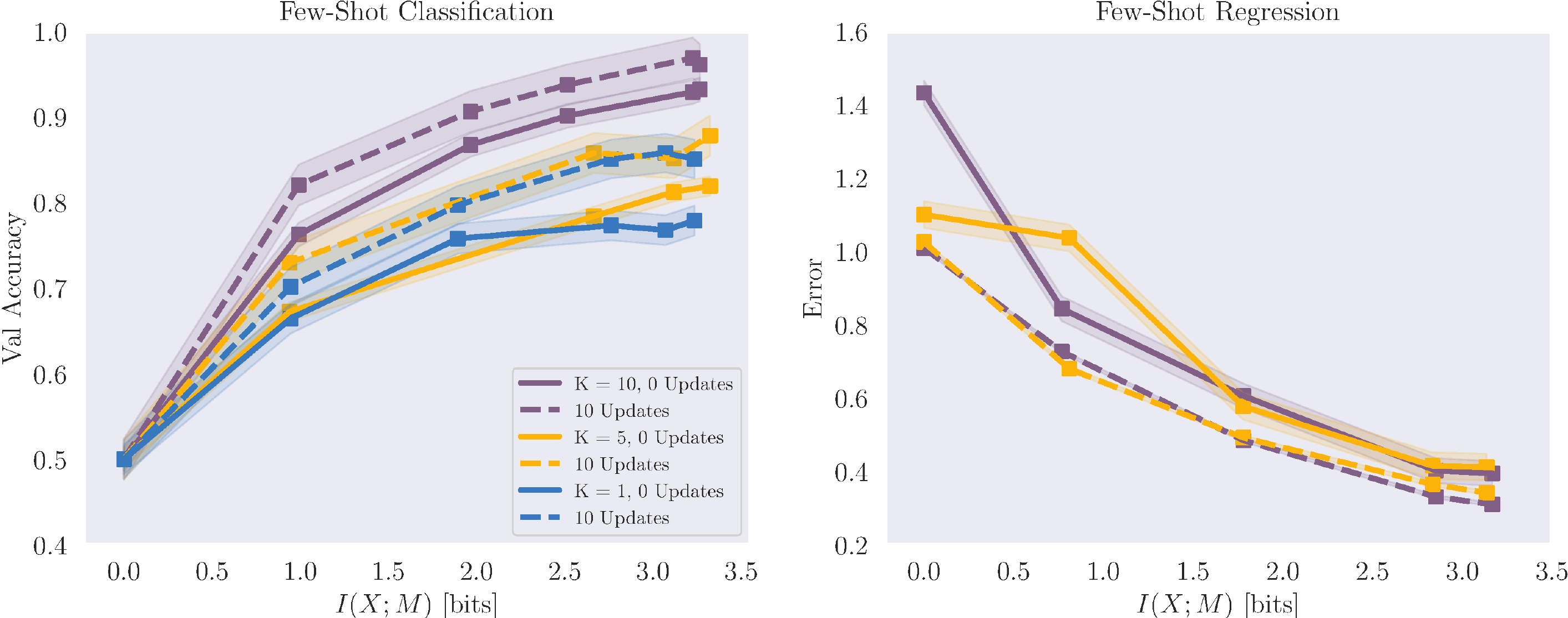}
\end{minipage}
\caption{The single expert system is not able to learn the structure of the sine wave, where the two expert can capture the periodic structure. The lower figure shows how the error decreases as we add more experts in the classification and regression setting, indicating successful specialization and expert selection.} 
\label{fig:sine}
\end{figure}

\section{Results}
\begin{figure*}[t!]
\begin{minipage}{\textwidth}
\centering
\includegraphics[width=\textwidth, trim={7cm 0cm 7cm 1cm}, clip]{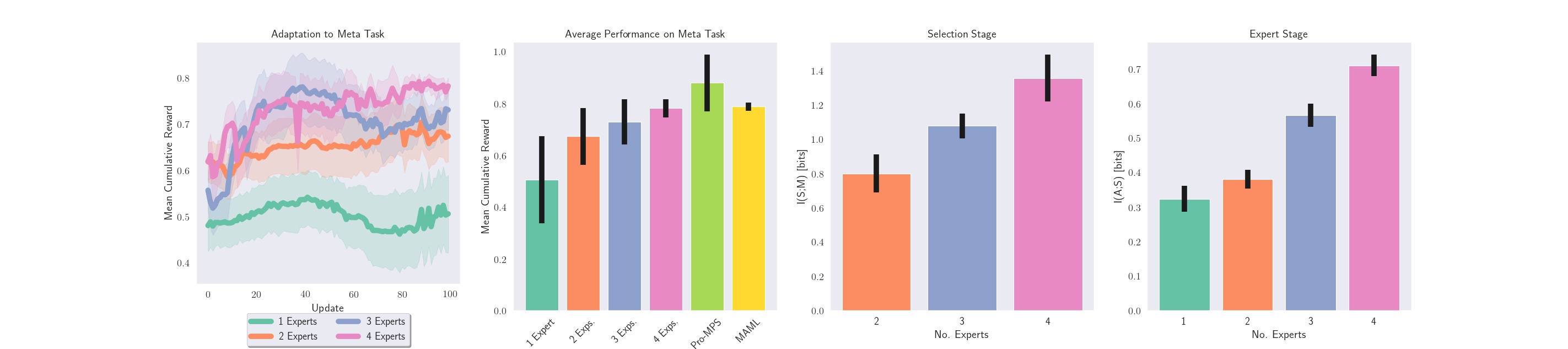}
\end{minipage}
\caption{In each Meta-Update Step we sample $N$ tasks from the training task set $T$ and update the agents. After training we evaluate their performance on a tasks from the meta test set $T'$. Rewards are normalized to $[0,1]$ and the episode horizon is 100 time steps. The agent achieves higher reward when adding more experts while the information-processing of the selection and of the expert stage increases, indicating that the added experts specialize successfully. We achieve comparable results to MAML \cite{Finn2017model} and Proximal Meta-Policy Search \cite{rothfuss2018promp}. Shaded areas and error bars represent one standard deviation. In Table \ref{tab:rlenvs} in the Appendix we give experimental details.}
\label{fig:metarl}
\end{figure*}
\subsection{Sinusoid Regression}
\label{sec:regressionresults}
\begin{figure}
\includegraphics[width=\linewidth, trim={0cm 6.375cm 0cm 0cm}, clip]{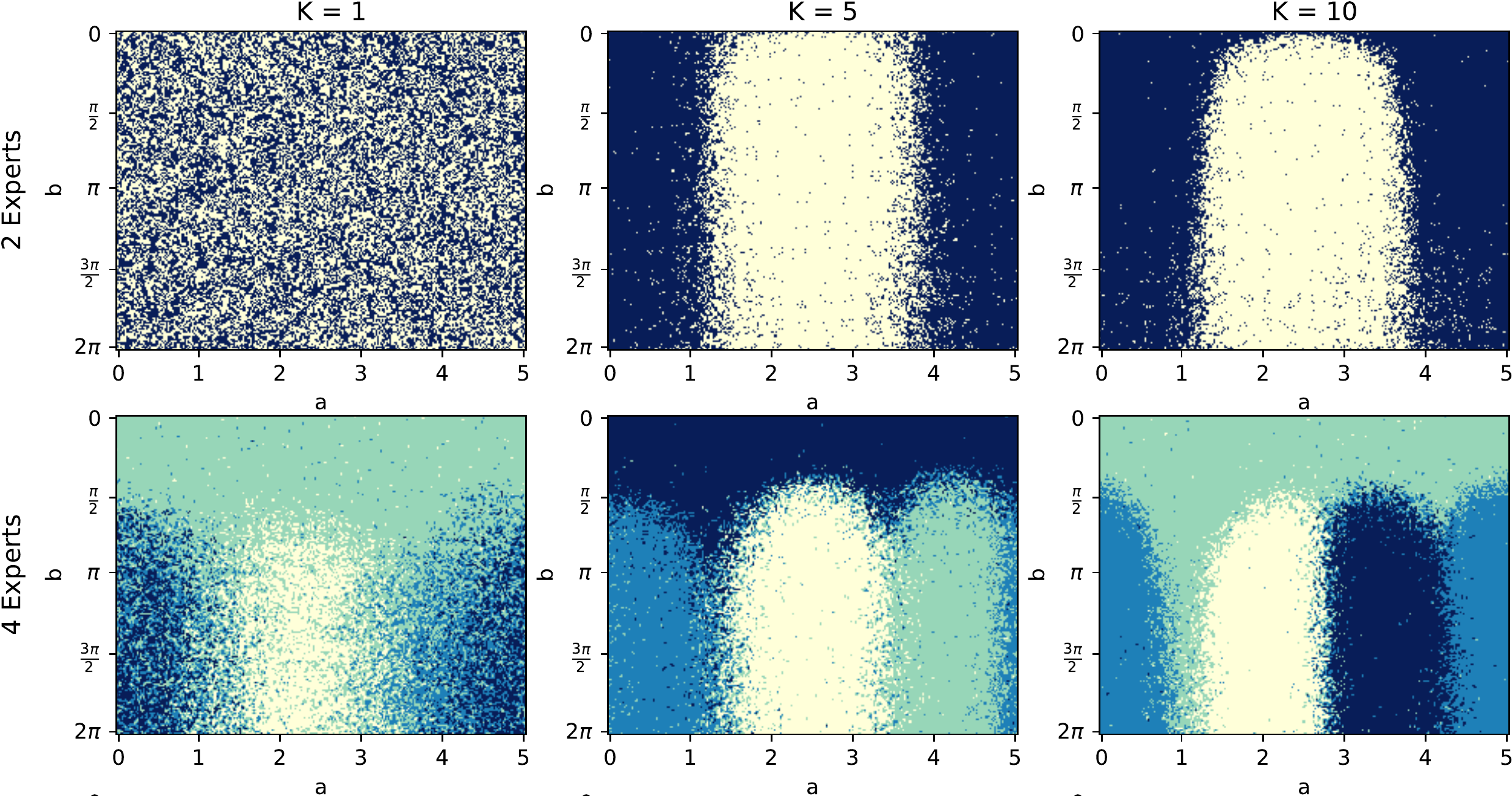}
\caption{Here we show the soft-partition found by the selection policy for the sine prediction problem, where each color represents an expert. We can see that the selection policy becomes increasingly more precise as we provide more points per dataset to the system.}
\label{fig:partition}
\end{figure}

We adopt this task from \cite{Finn2017model}. In this $K$-shot problem, each task consists of learning to predict a function of the form $y = a \cdot \sin(x + b)$, with both $a \in [0.1, 5]$ and $b \in [0, 2\pi]$ chosen uniformly, and the goal of the learner is to find $y$ given $x$ based on only $K$ pairs of $(x, y)$. Given that the underlying function changes in each iteration it is impossible to solve this problem with a single learner. Our results show that by combing expert networks, we are able to reduce the generalization error iteratively as we add more experts to our system--see Figure \ref{fig:sine} where we also show how the system is able to capture the underlying problem structure as we add more experts. In Figure \ref{fig:partition} we visualize how the selector's partition of the problem space looks like. The shape of the emerged clusters indicates that the selection is mainly based on the amplitude $a$ of the current sine function, indicating that from an adaptation point-of-view it is more efficient to group sine functions based on amplitude $a$ instead of phase $b$. We can also see that an expert specializes on the low values for $b$ as it covers the upper region of the $a\times b$ space. The selection network splits this region among multiple experts if we increase the set of experts to 8 or more.

\subsection{Few-Shot Classification}
The Omniglot dataset \cite{lake2011one} consists of over 1600 characters from 50 alphabets. As each character has only 20 samples each drawn by a different person, this forms a challenging meta-learning benchmark.

The selection policy now selects experts based on their free-energy that is computed over datasets $D_{\text{val}}$ and the selection policy depends on the training datasets $D_{\text{train}}$:
\begin{equation}
\max_\theta \E_{p_\theta(m\vert D_{\text{train}})}\left[\hat{f}(m,D_{\text{val}}) - \frac{1}{\beta_1}\log\frac{p_\theta(m\vert D_\text{train})}{p(m)}\right],
\end{equation}
where $\hat{f}(m,D_{\text{val}}) \coloneqq \mathbb{E}_{p_\vartheta(\hat{y}\vert m,x)}\big[-\mathcal{L}(\hat{y},y) - \frac{1}{\beta_2}\log\frac{p_\vartheta(\hat{y}\vert x,m)}{p(\hat{y}\vert m)}\big]$ is the free energy of expert $m$ on dataset $D_{\text{val}}$, $\mathcal{L}(\hat{y},y)$ is loss function, and $(x,y) \in D_\text{val}$. The experts optimize their free energy objective on the training dataset $D_{\text{train}}$ defined by
\begin{equation}
\max_\vartheta \E_{p_\vartheta(\hat{y}\vert m,x)}\left[-\mathcal{L}(\hat{y}, y) - \frac{1}{\beta_2}\log\frac{p_\vartheta(\hat{y}\vert x,m)}{p(\hat{y}\vert m)}\right],
\end{equation} 
where $(x, y) \in D_\text{train}$.


We train the learner on a subset of the dataset ($\approx 80\%$, i.e.,~ $\approx$ 1300 classes) and evaluate the remaining $\approx 300$ classes, thus investigating the generalization. In each training episode we build the datasets $D_{\text{train}}$ and $D_{\text{val}}$ by selecting a target class $c_t$ and sampling $K$ positive and $K$ negative samples. To generate negative samples, we draw $K$ images randomly out of the remaining $N-1$ classes. We present the selection network with the feature presentation of the $K$ positive training samples (see Figure \ref{fig:model}), but evaluate the experts' performance on the $2K$ test samples in $D_{\text{val}}$. In this way, the free energy of the experts becomes a generalization measure. Using this optimization scheme, we train the networks to become \textit{experts} in recognizing a subset of classes. After a proper expert is selected we train that expert using the $2K$ samples from the training dataset. 

We consider three experimental setups: 1) how does a learner with only a single hidden layer perform when trained n\"{i}vely compared to with sophisticated methods such as MAML \cite{Finn2017model} and Matching Nets \cite{vinyals2016matching} as a baseline? 2) does the system benefit from adding more experts and if so, at what rate? and 3) how does our method compare to the aforementioned algorithms? Regarding 1) we note that introducing constraints by reducing the representational power of the models does not facilitate specialization is it would by explicit information-processing constraints. In the bottom row of Figure \ref{fig:sine} we address question 2).  We can interpret this curve as the rate-utility curve showing the trade-off between information processing and expected utility (transparent area represents one standard deviation), where increasing $I(X;M)$ improves adaptation. The improvement gain grows logarithmically, which is consistent with what rate-distortion theory would suggest. In Table \ref{tab:classtable} we present empirical results addressing question 3).

\subsection{Meta Reinforcement Learning}
We create a set of RL tasks by sampling the parameters for the Inverted Double Pendulum problem \cite{Sutton1996} implemented in OpenAI Gym \cite{Brockman2016}, where the task is to balance a two-link pendulum in an upward position. We modify inertia, motor torques, reward function, goal position and invert the control signal -- see Table \ref{tab:rlenvs} for details. We build the meta task set $T'$ on the same environment, but change the parameter distribution and range, providing new but similar reinforcement learning problems. During training, we sample $M$ environments in each episode and perform updates as described earlier. During evaluation, we measure the system's performance on tasks sampled from $T'$-- see results in Figure \ref{fig:metarl}, where we can see improving performance as more experts are added and the increasing mutual information in the selection stage indicates precise partitioning.

The expert policies are trained on the meta-training environment policies, but evaluated on unseen but similar validation environments. In this setting we define the discounted free energy $\bar{F}_t$ with $$\bar{f}(x,m) = \mathbb{E}_{p_\vartheta(a\vert x,m)}\left[r_{\text{val}}(x,a) - \frac{1}{\beta_2} \log \frac{p(a\vert x, m)}{p(a\vert m)}\right],$$ where $r_{\text{val}}$ is a reward function defined by a validation environment (see Figure \ref{fig:metarl} for details).

\section{Related Work}
The hierarchical structure we employ is related to Mixture of Experts (MoE) models. \cite{Jacobs1991,Jordan1994} introduced MoE as tree structured models for complex classification and regression problems, where the underlying approach is a divide and conquer paradigm. As in our approach, three main building blocks define MoEs: gates, experts, and a probabilistic weighting to combine expert predictions. Learning proceeds by finding a soft partitioning of the input space and assigning partitions to experts performing well on the partition. In this setting, the model response is then a sum of the experts' outputs, weighted by how confident the gate is in the expert's opinion (see  \cite{Yuksel2012} for an overview). This paradigm has seen recent interest in the field of machine learning, see e.g.,~ \cite{aljundi2017expert,rosenbaum2019dispatched,jerfel2019reconciling,Shazeer2017}. The approach we propose allows learning such models, but also has applications to more general decision-making settings such as reinforcement learning. 

Our approach belongs to a wider class of models that use information constraints for regularization to deal more efficiently with learning and decision-making problems~\cite{Martius2013,Leibfried2017,Grau-Moya2017,Achiam2017,Hihn2018,grau2018soft}. One such prominent approach is Trust Region Policy Optimization (TRPO) \cite{Schulman2015}. The main idea is to constrain each update step to a trust region around the current state of the system. This region is defined by $\DKL(\pi_\text{new}||\pi_\text{old})$ between the old policy and the new policy, providing a theoretic monotonic policy improvement guarantee. In our approach we define this region by $\DKL$ between the agent's posterior and prior policy, thus allowing to learn this region and to adapt it over time.  This basic idea has been extend to meta-learning by \cite{rothfuss2018promp}, which we use to compare our method against in meta-rl experiments. 

Most other methods for meta-learning such as the work of \cite{Finn2017model} and \cite{ravi2017optimization}  find an initial parametrization of a single learner, such that the agent can adapt quickly to new problems. This initialization represents prior knowledge and can be regarded as an abstraction over related tasks and our method takes this idea one step further by finding a possibly disjunct set of such compressed task properties. Another way of thinking of such abstractions by lossy compression is to go from a task-specific posterior to a task-agnostic prior strategy. By having a set of priors the task specific information is available more locally then with a single prior, as in MAML \cite{Finn2017model} and the work of \cite{ravi2017optimization}. In principle, this can help to adapt within fewer iterations. Thus our method can be seen as the general case of such monolithic meta-learning algorithms. Instead of learning similarities within a problem, we can also try to learn similarities between different problems (e.g.,~ different classification datasets), as is described in the work of \cite{yao2019hierarchically}. In this way, the partitioning is governed by different tasks, where our study however focuses on discovering meta-information within the same task family, where the meta-partitioning is determined solely by the optimization process and can thus potentially discover unknown dynamics and relations within a task family. 

\section{Discussion}
We have introduced and evaluated a novel hierarchical information-theoretic approach to meta-learning. In particular, we leveraged an information-theoretic approach to bounded rationality. 
Although our method is widely applicable, it suffers from low sample efficiency in the RL domain. A research direction would be to combine our system with model-based RL which is known to improve sample efficiency. Another research direction would be to investigate the performance of our system in continual adaption tasks, such as in \cite{yao2019hierarchically}. Another limitation is the restriction to binary meta classification tasks, which we leave for feature work.

The results show that our method can identify sub-regions of the problem set and solve them efficiently with expert networks. In effect, this equips the system with  initializations covering the problem space and thus enables it to adapt quickly to new but similar tasks. To reliably identify such tasks, we have proposed feature extraction methods for classification, regression, and reinforcement learning, that could be simply be replaced and improved in future work. The strength of our model is that it follows from simple principles that can be applied to a large range of problems. Moreover, the system performance can be interpreted in terms of the information processing of the selection stage and the expert decision-makers. We have shown empirically that our method achieves results comparable to recent meta-learning algorithms, such as MAML \cite{Finn2017model}, Matching Nets \cite{vinyals2016matching}, and Proximal Meta-Policy Search \cite{rothfuss2018promp}.

\subsubsection*{Acknowledgments}
This work was supported by the European Research Council Starting Grant \emph{BRISC}, ERC-STG-2015, Project ID 678082.

\bibliographystyle{icml2020}
\bibliography{bibliography.bib}

\section*{Appendix}
\textbf{Few-Shot Classification Experimental details:} We followed the design of \cite{vinyals2016matching} but reduce the number of blocks to one. We used a single convolutional block consisting of 32 3$\times$3 filters with strided convolutions followed by a batch normalization layer and a ReLu non-linearity. During training we used a meta-batch size of 16. The convolutional autoencoder is a 3 layer network consisting of 16, 16, and 4 filters each with size 3$\times$3 with strided convolutions followed by a leaky ReLu non-linearity. The layers are mirrored by de-convolotional layers to reconstruct the image. This results in an image embedding with dimensionality 64. The selection network is a two layer network with 32 units each, followed by a ReLu non-linearity, a dropout layer \cite{srivastava2014dropout} per layer. We augment the dataset by rotating each image in 90, 180, and 270 degrees resulting in 80 images per class. We also normalize the images two be in (0,1) range. We evaluate our method by resetting the system to the state after training and allow for 10 gradient updates and report the final accuracy. To evaluate MAML on 2-way $N$-shot omniglot dataset we used a inner learning rate of $\alpha = 0.05$ and one inner update step per iteration for all settings. We used a single convolutional block followed by a fully connected layer with 64 units and a ReLU non-linearity. Note, that we reduce the number of layers to make the tests comparable. Using the suggested architectures by \cite{Finn2017model} we achieve classification accuracy $\geq 95\%$.  To generate this figure, we ran 10-fold cross-validation on the whole dataset and show the averaged performance metric and the standard-deviation across the folds. In both settings, ''0 bits'' corresponds to a single expert, i.e.,~ a single neural network trained on the task. 

\textbf{Meta-RL Experimental details:} The selector's actor and critic net are build of RNNs with 200 hidden units each. The critic  is trained to minimize the Huber loss between the prediction and the cumulative reward. The experts are two layer networks with 64 units each followed by ReLu non-linearities and used to learn the parameters of a Gaussian distribution. The critics have the same architecture (except for the output dimensionality). The control signal $a$ is continuous in the interval [-1,1] and is generated by neural network that outputs $\mu$ and $\log(\sigma)$ of a gaussian. The action is then sampled by re-parameterizing the distribution to $p(a) = \mu + \exp(\sigma)\epsilon$, where $\epsilon \thicksim \mathcal{N}(0,1)$, so that the distribution is differentiable w.r.t to the network outputs.  We average the results over 10 random seeds and trained for 1000 episodes each with a batch of 64 environments. 

\begin{table}[H]
\centering
\begin{tabular}{lcc}
\toprule
& \multicolumn{2}{c}{\textbf{Task Distribution}}\\
\textbf{Paramater} & \textbf{$T$} & \textbf{$T'$}  \\
\midrule
Distance Penalty & [$10^{-3},10^{-1}$] & [$10^{-3},10^{-2}$] \\
Goal Position & [0.3, 0.4] & [0, 3]\\
Start Position & [-0.15, 0.15] & [-0.25, 0.25] \\
Motor Torques & $[0, 5]$ & $[0, 3]$\\
Inverted Control & $p = 0.5$ & $p = 0.5$  \\
Gravity & [0.01, 4.9] & [4.9, 9.8] \\
Motor Actuation & [185, 215] & [175, 225] \\
\bottomrule
\end{tabular}
\caption{Environment Parameters for the Meta-RL Setting.}
\label{tab:rlenvs}
\end{table}

\textbf{Regression Experimental details:} For regression we use a two layer selection network with 16 units each followed by tanh non-linearities. The experts are shallow neural networks with a single hidden layer that learn log-variance and mean of a Gaussian distribution which they use for prediction. We use the ``Huber Loss" instead of MSE as it is more robust \cite{balasundaram2019robust}. We optimize all networks using Adam \cite{kingma2014adam}.

\end{document}